\documentclass[preprint,12pt]{elsarticle}

\usepackage{amssymb}
\usepackage{amsmath,amsfonts}
\usepackage{algorithmic}
\usepackage{algorithm}
\usepackage{graphicx}
\usepackage{amsmath}
\usepackage{amsthm}
\usepackage{amssymb}
\usepackage{booktabs}
\usepackage{bbm}
\usepackage{color}
\usepackage{makecell}
\usepackage{footnote}
\usepackage{threeparttable}
\usepackage{wrapfig}

\usepackage{array}
\usepackage{textcomp}
\usepackage{stfloats}
\usepackage{url}
\usepackage{verbatim}
\usepackage{graphicx}
\usepackage{cite}
\usepackage{multirow}
\usepackage{float}
\usepackage[table]{xcolor} 

\journal{Information Sciences}

\begin{document}
\include{preamble}

\begin{frontmatter}




\title{Mitigating Spurious Correlations with Causal \\ Logit Perturbation\tnoteref{t1}}

\author[1]{Xiaoling Zhou}

\author[1]{Wei Ye\corref{mycorrespondingauthor}}
\cortext[mycorrespondingauthor]{Corresponding author}
\ead{wye@pku.edu.cn}

\author[1]{Rui Xie}

\author[1]{Shikun Zhang\corref{mycorrespondingauthor}}
\ead{zhangsk@pku.edu.cn}

\affiliation[1]{organization={National Engineering Research Center for Software Engineering},
            addressline={Peking University},
            city={Beijing},
            postcode={100871},
            country={China}}



\begin{abstract}
Deep learning has seen widespread success in various domains such as science, industry, and society. However, it is acknowledged that certain approaches suffer from non-robustness, relying on spurious correlations for predictions. Addressing these limitations is of paramount importance, necessitating the development of methods that can disentangle spurious correlations. {This study attempts to implement causal models via logit perturbations and introduces a novel Causal Logit Perturbation (CLP) framework to train classifiers with generated causal logit perturbations for individual samples, thereby mitigating the spurious associations between non-causal attributes (i.e., image backgrounds) and classes.} {Our framework employs a} perturbation network to generate sample-wise logit perturbations using a series of training characteristics of samples as inputs. The whole framework is optimized by an online meta-learning-based learning algorithm and leverages human causal knowledge by augmenting metadata in both counterfactual and factual manners. Empirical evaluations on four typical biased learning scenarios, including long-tail learning, noisy label learning, generalized long-tail learning, and subpopulation shift learning, demonstrate that CLP consistently achieves state-of-the-art performance. Moreover, visualization results support the effectiveness of the generated causal perturbations in redirecting model attention towards causal image attributes and dismantling spurious associations.
\end{abstract}




\end{frontmatter}


\section{Introduction}

\begin{figure}[t] 
\centering
\includegraphics[width=0.95\textwidth]{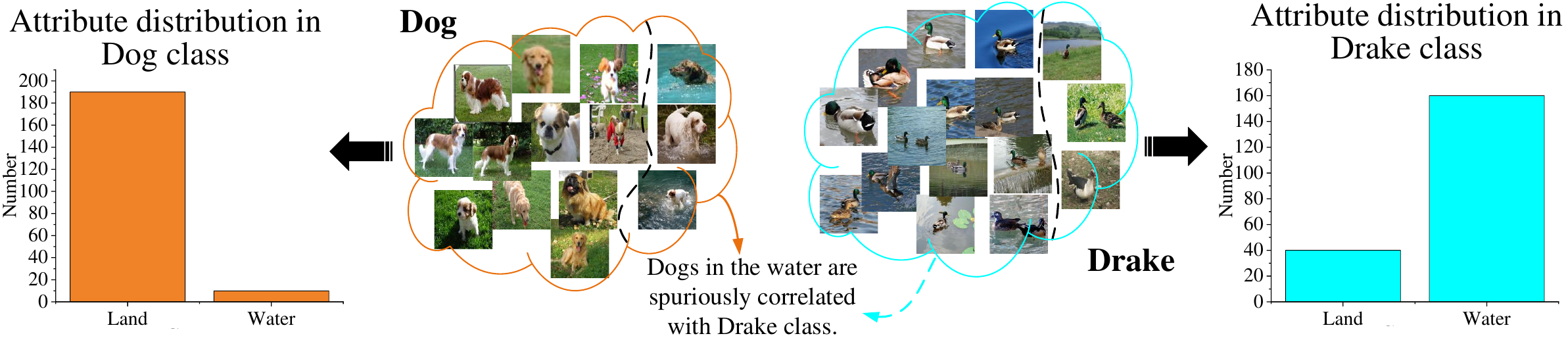}
\caption{{Illustration for spurious correlation resulting from a rare background occurrence. Dogs depicted in water are easy to be misclassified as drakes, primarily due to the infrequency of water backgrounds in dog class.}}
\label{fig11}
\end{figure}

The direct minimization of empirical risk may result in the learning of unstable spurious correlations, which overlook the underlying causal structures~\citep{r58}. Particularly, in high-dimensional feature spaces characterized by strong correlations, identifying appropriate feature sets that embody causal associations for accurate target prediction becomes challenging~\citep{r59,boostingzxl}. {This issue arises because different feature sets can yield identical training accuracy, thereby promoting spurious associations between attributes and labels that lack genuine causal foundations~\citep{r60}.} For example, the background of an image, although seemingly irrelevant to its foreground labeling, can spuriously correlate with labels. Fig.~\ref{fig11} illustrates a scenario where, due to the prevalence of grassland backgrounds for most dogs in the dog class, dogs in water may be mistakenly classified as drakes. {When the distributions of the training and test data are inconsistent, if the model overfits spurious correlations present in the training data that may be absent in the test set, it will perform well on the training set but poorly on the test set.}

{To enhance model performance, a promising avenue involves the pursuit of causal representations that utilize the causal components of instances for decision-making~{\citep{DORAL}}.} Achieving robust generalization hinges on the capacity of such representations to discern the causal mechanisms governing image features and category labels. However, conventional causal inference methods such as randomized control trials or interventions are impractical for passively collected natural images due to the inability to control or manipulate variables. Consequently, researchers have turned to counterfactual data augmentation as a feasible alternative, given its ability to break spurious correlations and its computational, psychological, and legal benefits. For instance, Sauer and Geiger~\citep{sauer95} proposed a method for generating counterfactuals using a learned generative model that effectively disentangles foreground and background attributes. Additionally, Chang et al.~\citep{saldefine} introduced two image generation procedures, incorporating counterfactual and factual augmentations, well mitigating spuriousness.

It is worth noting that logit perturbation techniques have proven to be a valuable approach for improving the generalization of models. However, the perturbation terms in existing methods typically focus on only one or two factors, such as class proportion and sample difficulty. Additionally, these terms are often artificially defined, limiting their applicability to specific learning scenarios, such as those involving class imbalances. For example, LA~\citep{LA} was introduced for long-tailed classification, wherein it decreases losses for samples in the head categories while increasing those in the tail categories, thereby bolstering model performance on tail classes. Similarly, LDAM~\citep{LDAM} was designed to tackle imbalance learning by augmenting the margins of samples in tail categories. Furthermore, LPL~\citep{LPL} adopts distinct perturbation strategies for head and tail categories, applying positive or negative augmentation based on loss maximization and minimization directions in a heuristic manner. 
While these methods have achieved success in specific learning scenarios, their reliance on a limited number of factors and their category-level granularity impose constraints on their adaptability across diverse learning contexts. {Furthermore, these methods fail to transfer causal knowledge to the model and are incapable of eliminating spurious associations within models.}

{This study is the first to investigate the construction of causal models via logit perturbation.} Specifically, we aim to generate causal logit perturbations for dismantling spurious correlations between non-causal attributes (i.e., image backgrounds) and labels, thereby enhancing the generalization and robustness of deep classifiers. 
A novel {logit perturbation} framework, termed {C}ausal {L}ogit {P}erturbation (CLP), is proposed, which consists of {four main} modules: the metadata augmentation module, the backbone classifier, the training characteristics extraction module, and the perturbation network which generates logit perturbations of samples. {To optimize the entire framework, a meta-learning-based strategy is employed to alternately update the classifier and the perturbation network, where the utilized metadata are augmented in two ways: counterfactual and factual augmentations, to incorporate human causal knowledge into the perturbation generation process.}
Additionally, {a series of training characteristics, such as training loss, margin, and uncertainty, are extracted for each sample and fed into the perturbation network to guide the generation of the perturbations.}
Consequently, unlike conventional logit perturbation techniques that rely on explicit perturbation functions, CLP devises perturbations by leveraging a trained perturbation network (represented by a two-layer MLP) based on the training characteristics of samples. 

To validate the efficacy of our approach, we conduct extensive evaluations {across} four typical learning scenarios {where spurious correlations are prone to occur}: long-tail {(LT)} learning, noisy label learning, subpopulation shifts learning, and generalized LT {(GLT)} learning. The experimental results demonstrate that CLP consistently achieves state-of-the-art (SOTA) performance across all learning scenarios. Moreover, through our visualization analyses, we provide compelling evidence that the generated perturbations effectively disrupt spurious correlations within the models, {encouraging the model to prioritize causal elements in the images}.
{Overall,} the contributions of {our} study can be summarized as follows:

\begin{itemize}
    \item We present a novel logit perturbation framework, CLP, which trains classifiers with causal logit perturbations, thereby effectively mitigating spurious correlations and enhancing {model generalization}. Remarkably, CLP represents a pioneering effort in achieving causal models through logit perturbation.
    \item We introduce both counterfactual and factual data augmentation manners to augment the meta dataset, which effectively incorporates causal knowledge into our proposed framework and guides the training of the perturbation network. 
    \item Our proposed CLP framework undergoes rigorous evaluation in four learning scenarios that exhibit susceptibility to spurious correlations. The comprehensive evaluation substantiates the efficacy of CLP in effectively disrupting spurious correlations and enhancing the generalization performance of classifiers.
\end{itemize}

\section{Related Work}


\paragraph{Causal Inference} {Causal inference~\citep{review,valuaingzxl} has been extensively adopted across various disciplines, including psychology, political science, and epidemiology. It functions both as an interpretive framework and as a tool for achieving specific goals by investigating causal effects. {Recently}, its application has expanded into the field of computer vision, where it is used to address issues related to bias~\citep{CNC}.} {Among the existing causal inference techniques}, counterfactual augmentation has emerged as a prominently investigated one due to its advantages in computation and psychology. {This approach involves augmenting training data} with counterfactual modifications to a subset of the causal factors, which leaves the rest of the factors untouched. {For example, Singla et al.~\citep{ACE} introduced ACE, a method that fine-tunes a pre-trained classifier using augmentations derived from a counterfactual explainer.}
{Besides, Deng et al.~\citep{PDE} proposed PDE to enhance model robustness and improve performance for the worst-performing group.}
In addition, 
Mao et al.~\citep{genint} leveraged a causal-agnostic generative model to generate counterfactual samples. 
{In contrast to previous studies that focus on augmenting training data, our approach applies counterfactual and factual augmentations to the meta data, which is introduced through meta-learning to guide the training of the perturbation network for generating logit perturbations. This enables the perturbation network to generate logit perturbations that assist the model in breaking spurious correlations and enhancing its ability to capture causal relationships.}

\paragraph{Logit Perturbation} Logit perturbation technique has proven to be effective across a range of deep learning scenarios~{\citep{lp1,classzxl}}. In these approaches, the logits, representing the outputs of the final feature encoding layer in most deep neural networks, undergo perturbation practices, primarily at the class level. 
For instance, LPL~\citep{LPL} addresses data imbalance by employing distinct perturbation methods for head and tail categories. On the other hand, {GCL~\citep{lp1} introduces feature augmentation to balance the embedding distribution by perturbing the features of different classes with varying amplitudes in Gaussian form.} 
{Moreover,} the ALA loss~\citep{ALA} accounts for two factors including class proportion and learning difficulty of samples to effectively handle data imbalance. Additionally, several implicit data augmentation methods actually perturb the logits of samples~\citep{RISDA,implicitzxl}.
However, the perturbation terms in these methods are always artificially defined and only rely on one or two factors, limiting their effectiveness across various learning scenarios, such as noisy learning environments. {Unlike prior methods, our approach employs a perturbation network that dynamically and adaptively generates perturbations during training, guided by a range of sample-level training characteristics, thereby making it well-suited to diverse learning scenarios.}

%


\paragraph{Meta-Learning} 
{In recent years, meta-learning has garnered considerable attention within the research community~\citep{MAML}. These methods can be classified into three categories: metric-based, model-based, and optimization-based techniques~\citep{metasaug}.} Nowadays, an algorithm from the optimizing-based methods category, drawing inspiration from Model-Agnostic Meta-Learning~\citep{MAML} has gained increasing attention. In this paradigm, a machine learning model assimilates knowledge from multiple learning episodes and employs it to enhance its learning performance. {The meta-optimization process is inherently data-driven. Existing research typically involves learning sample weights or parameters from an additional meta dataset.} For example, Meta-Weight-Net~\citep{shu2019meta} utilizes this learning algorithm to alternatively update the parameters in the classifier and the weighting network, which has shown good performance when dealing with biased datasets. 
Our method also belongs to this paradigm, enabling the alternative updating of the classification network and the perturbation network.

\section{Methodology}

\paragraph{Notation.}
Let $\boldsymbol{x}\in\mathcal{X}$ denote the sample instance and $y\in\mathcal{Y}$ denote the label, where $\mathcal{X}\subseteq\mathbb{R}^{d}$ represents the instance space, {with $d$ denoting the dimensionality of the input features}, and $\mathcal{Y}=\{1,\cdots,\mathcal{C}\}$ denotes the label space, with $\mathcal{C}$ referring to the number of classes. The classification model $\mathcal{F}$ maps the input data space $\mathcal{X}$ to the label space $\mathcal{Y}$. The training set is denoted as $\{(\boldsymbol{x}_{i},y_{i})\}_{i=1}^{\mathcal{N}}$, with $\mathcal{N}$ refers to the number of training samples. The logit vector of $\boldsymbol{x}$ is represented as $\boldsymbol{u}_{\boldsymbol{x}}=\mathcal{F}(\boldsymbol{x})\in \mathbbm{R}^{\mathcal{C}}$, and its logit perturbation term is denoted as $\boldsymbol{\delta}_{\boldsymbol{x}}$.

\begin{figure}[t] 
\centering
\includegraphics[width=0.95\textwidth]{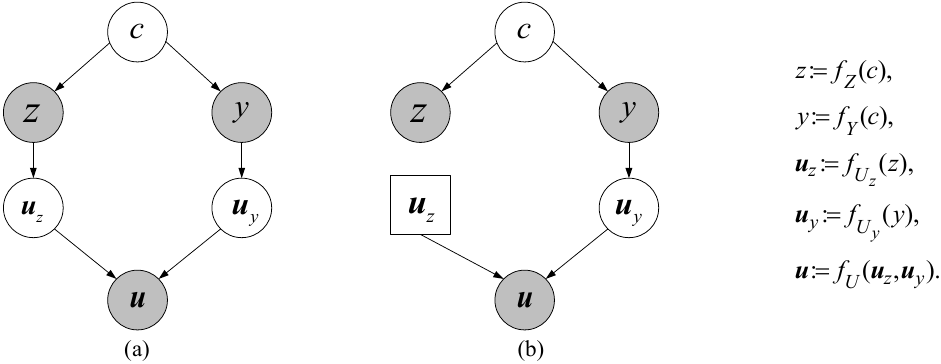}
\caption{The causal graph before (a) and after (b) the intervention process of $\boldsymbol{u}_{z}$. By intervening on $\boldsymbol{u}_{z}$, the backdoor path $z\!\leftarrow\! c\!\rightarrow\!y$ is broken. The observed and unobserved variables are represented by white and grey nodes, respectively, while the intervened variable is depicted using a square. The variables follow the formulas on the right.}
\label{causal}
\end{figure}

\subsection{Causal Graph}
Fig.~\ref{causal}(a) shows the causal graph before {logit} perturbation, where $z$ {denotes} the non-causal attributes, $y$ represents the target, and $c$ represents the confounder. The vectors $\boldsymbol{u}_z$ and $\boldsymbol{u}_y$ correspond to the extrinsic and intrinsic components of the logit vector $\boldsymbol{u}$, respectively, that are caused by $z$ and $y$. 

The presence of the confounder $c$ creates a backdoor path $z \leftarrow c \rightarrow y$, thereby causing a spurious correlation between $z$ and $y$. To eliminate this issue, we intervene in $\boldsymbol{u}_{z}$, represented as $\text{do}(\boldsymbol{u}_{z})$, to close this backdoor path. Fig.~\ref{causal}(b) illustrates the causal graph after intervention on $\boldsymbol{u}_{z}$, in which the backdoor path caused by $c$ is broken. {Therefore, our objective is to design the perturbation function $p$ for the logit vector $\boldsymbol{u}$
that mimics an intervention on $\boldsymbol{u}_{z}$:}
\begin{equation}
    \tilde{\boldsymbol{u}} = p(\boldsymbol{u}) = f_{U}(do(\boldsymbol{u}_{z}),\boldsymbol{u}_{y}),
\end{equation}
{where {$\tilde{\boldsymbol{u}}=p(\boldsymbol{u})$} refers to the perturbed logit vector. 
In practical applications, our approach ensures that logit perturbations, which contribute to breaking spurious correlations in the model, are generated by incorporating human causal knowledge. {The process by which the logit perturbations are generated through the incorporation of human causal knowledge will be further elaborated upon in the subsequent subsection.}}

{In summary, the objective of our logit perturbation operation is to intervene on the extrinsic components of the logit vector $\boldsymbol{u}_{z}$, enabling the model, when trained with the perturbed logit vectors, to disrupt backdoor paths and mitigate its reliance on spurious associations. This is made possible because the perturbation network is trained on a meta dataset that integrates human causal knowledge, which allows the generated logit perturbations to effectively achieve the desired outcome.}

\begin{figure*}[t] 
\centering
\includegraphics[width=1\textwidth]{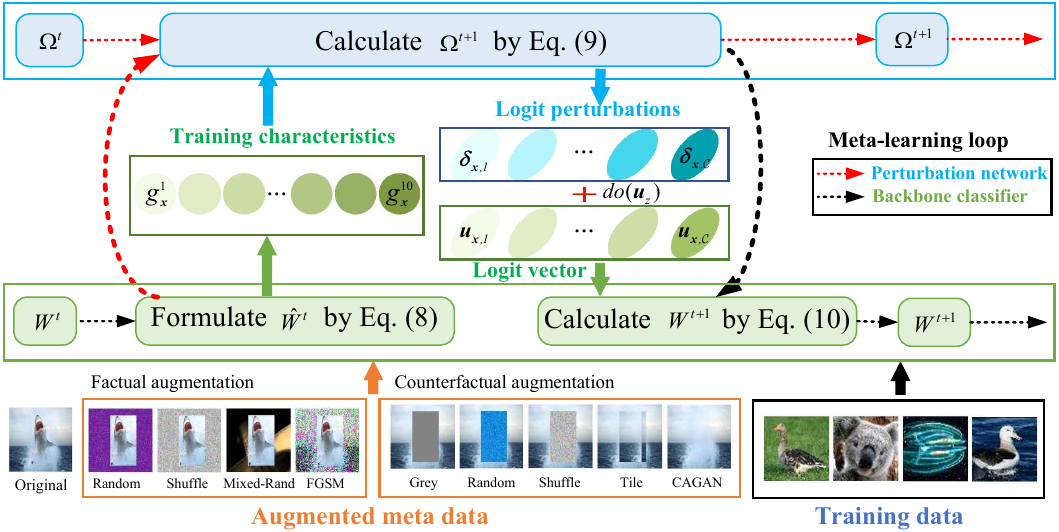}
\caption{{The overall structure of CLP, {which consists of four main components: the metadata augmentation module, the backbone classifier, the training characteristics module, and the perturbation network which generates sample-wise logit perturbations.} The black and red lines indicate the learning loops of the backbone classifier and the perturbation network, respectively.}}
\label{frame}
\end{figure*}

\subsection{Causal Logit Perturbation Framework}
The proposed CLP framework mitigates spurious correlations between class labels and image backgrounds in deep classifiers by perturbing the sample logits. 
Fig.~\ref{frame} illustrates the {whole} structure of the CLP framework, which consists of four key components: the metadata augmentation module, the backbone classifier, the training characteristics extraction module, and the perturbation network which generates sample-wise logit perturbations. The optimization of the entire framework employs a meta-learning approach, alternating between optimizing the classifier and the perturbation network. To mitigate spurious associations between image backgrounds and class labels, our approach leverages human causal knowledge into the meta dataset. Specifically, the meta dataset undergoes both counterfactual and factual augmentations, {which can then guide the training of the perturbation network}. A series of training characteristics of samples are extracted from the classifier and input into the perturbation network to {facilitate} generating sample-specific {logit} perturbations. {Consequently, utilizing the perturbation terms generated by the perturbation network, which is optimized on the causally-augmented metadata, will facilitate the classifier to behave well on the metadata, thereby helping break spurious correlations.} Notably, our framework exhibits the following advantages over existing relevant studies:
\begin{itemize}
\item CLP generates sample-wise perturbations through a perturbation network that considers a comprehensive set of training characteristics of samples while existing techniques typically rely on heuristic definitions and {only consider} one or two factors.
\item CLP mitigates spurious correlations existing between image backgrounds and labels, a critical aspect often disregarded by conventional logit perturbation approaches.
\end{itemize}

Subsequent sections will provide detailed explanations of the metadata augmentation module, the training characteristic extraction module, and the meta-learning-based training algorithm.

\subsubsection{{Metadata Augmentation}}
{Guided by a small amount of metadata, the
parameters of the perturbation function can be finely updated simultaneously with
the learning process of the classifiers. {The small meta dataset, characterized by its cleanliness and balance, can be directly extracted from either the validation set or the training set.}} 
To assist the classifier in breaking spurious correlations between non-causal attributes (i.e., image backgrounds) and classes, we incorporate human causal knowledge into metadata. {Specifically, we humans know that the background of an image should not be correlated with its label.} {Accordingly,} the meta dataset is augmented in two ways: counterfactual and factual augmentations. {Counterfactual augmentation involves removing the causal components (i.e., foregrounds) while retaining the non-causal sections (i.e., backgrounds) of the image.} {Due to the alteration of the foregrounds, this type of augmentation changes the image label.} Counterfactuals can help break the spurious correlations between background and labels. Conversely, factual augmentation keeps the causal parts of the images intact while removing their backgrounds. Unlike counterfactual augmentation, this augmentation does not impact the image labels. Factual samples can help protect the classifier against background shifts. {By utilizing the augmented metadata to update the perturbation network, the generated logit perturbations prove beneficial in breaking spurious correlations between image backgrounds and labels.} 
{Take an image $\boldsymbol{x}^{{meta}}$ in metadata, comprising $G$ pixels, a label $y$, and a causal area $\boldsymbol{r}\in\{0,1\}^{G}$ (where $1$ represents causal),} {we employ different infilling methods to modify the causal and non-causal components of images.}


Counterfactual augmentation incorporates an infilling function $\Phi_{{cf}}$ to combine an original image with an infilling value $\hat{\boldsymbol{x}}^{{meta}}$:
\begin{equation}
\begin{aligned}
    \Phi_{{cf}}(\boldsymbol{x}^{{meta}}, \boldsymbol{r})&=(\boldsymbol{1}-\boldsymbol{r}) \odot \boldsymbol{x}^{{meta}}+\boldsymbol{r} \odot \hat{\boldsymbol{x}}^{{meta}}, 
\end{aligned}
\end{equation}
where $\odot$ denotes the Hadamard product. {These augmented images} are labeled as ``non-y" ($\neg y$). {The following are five simple methods to generate counterfactual infilling values. 1) \textbf{Grey} counterfactuals set all {infilling} pixels to 0.5. 2) \textbf{Random} counterfactuals begin by sampling from a uniform distribution, creating low-frequency noise. They then introduce high-frequency noise by adding Gaussians with $\sigma\!=\!0.2$ per channel per pixel. The resulting pixel values are then truncated to the range $[0,1]$. 3) \textbf{Shuffle} counterfactuals involve a random shuffling of pixels within the specified region, preserving the marginal distribution while disrupting the joint distribution. 4) \textbf{Tile} counterfactuals begin by extracting the largest rectangle from the background that does not overlap with the foreground. This rectangle is then used to tile the foreground region. 5) Contextual Attention GAN (\textbf{CAGAN}), an ImageNet model pre-trained for inpainting, is also included in the counterfactual generation process.}
We then devise a loss function for the counterfactually augmented samples. The negative log-likelihood of $P(\hat{y}\neq y)$ is employed, which is computed as
\begin{equation}
    \ell^{{CF}} = -\log\left(1-P\left(\hat{y}^{{cf}}=y^{{meta}}|\Phi_{{cf}}(\boldsymbol{x}^{{meta}})\right)\right),
\end{equation}
where $\hat{y}^{{cf}}$ is the prediction of $\Phi_{{cf}}(\boldsymbol{x}^{{meta}})$. Moreover, $y^{{meta}}$ is the label for $\boldsymbol{x}^{{meta}}$. 

The infilling function for factual augmentation ($\Phi_{{f}}$) that
mixes the foreground with the background is
\begin{equation}
\begin{aligned}
    \Phi_{{f}}(\boldsymbol{x}^{{meta}}, \boldsymbol{r})&=\boldsymbol{r} \odot \boldsymbol{x}^{{meta}}+(\boldsymbol{1}-\boldsymbol{r}) \odot \hat{\boldsymbol{x}}^{{meta}},
\end{aligned}
\end{equation}
In addition to employing existing counterfactual infilling methods like {Random} and {Shuffle}, we adopt the following two techniques to generate infilling values for factual samples. 1) The \textbf{Mix-Rand} method involves randomly swapping an image's background with the background of another tiled image randomly selected from different classes within the same training batch. 2) Manipulating non-causal features required the application of targeted adversarial attacks that specifically focused on the background area. {To facilitate the use of targeted adversarial attacks to manipulate non-causal features, the $\ell_{\infty}$ norm is used along with the fast-computing \textbf{FGSM} attack. Accordingly, during the augmentation process, the foreground of the image remains unchanged, while the background pixels are perturbed using the FGSM algorithm.} The lower portion of Fig.~\ref{frame} depicts the augmented counterfactual and factual examples. Throughout the subsequent sections, we refer to the counterfactual and factual augmentation methods as CF and F, respectively.

{Although the above method for generating fill values is effective,} many datasets encounter the absence of foreground segmentation masks, compelling us to adopt the HAttMatting method proposed in~\citep{HAttMatting}. This approach effectively discriminates between foreground and background pixels, enabling the prediction of alpha mattes solely from single RGB images. However, as the shape of the foreground still imparts information to the model, we utilize rectangular bounding boxes derived from the matting images, representing the smallest enclosing rectangles encompassing the foreground objects.

\subsubsection{{Training Characteristics Extraction}}
{Drawing on existing logit adjustment approaches~\citep{combinezxl}, logit perturbation terms for samples should be determined based on their training dynamics}.
{Therefore, we extract ten characteristics from the classification network that have been validated by previous research as effectively reflecting the training dynamics of samples while being computationally efficient:} $g_{{\boldsymbol{x}}}^{1}, \cdots, g_{{\boldsymbol{x}}}^{10}$, as depicted in the characteristic extraction module in Fig.~\ref{frame}. Then, these ten extracted training characteristics are input into the perturbation network to generate the logit perturbations of the samples.

\begin{table}[t]
\centering
\resizebox{1\textwidth}{!}{
\begin{tabular}{c|l|l}
\toprule
Index &Training characteristics & Formula \\ \midrule \midrule
1 & Sample loss~{\citep{shu2019meta}}  & $ g_{\boldsymbol{x}}^{1} = \ell_{\boldsymbol{x}}$  \\
2 & Sample margin~{\citep{LDAM}}  & $g_{\boldsymbol{x}}^{2} = \mathbbm{S}(\boldsymbol{u}_{\boldsymbol{x}})_{{y}_{\boldsymbol{x}}}-\text{max}_{j\neq {y}_{\boldsymbol{x}}}\mathbbm{S}(\boldsymbol{u}_{\boldsymbol{x}})_{j}$\\
3 & The norm of the loss gradient  & $g_{\boldsymbol{x}}^{3} = ||\boldsymbol{y}_{\boldsymbol{x}}-\mathbbm{S}(\boldsymbol{u}_{\boldsymbol{x}})||_{2}$ \\
4 &  Cosine value of the angle between features and weights~{\citep{ALA}}  & $g_{\boldsymbol{x}}^{4} = \cos\theta_{\boldsymbol{x},y_{\boldsymbol{x}}}$ \\ 

5 & The information entropy of the Softmax output~{\citep{Focal}}  & $g_{\boldsymbol{x}}^{5} = -\sum_{j=1}^{\mathcal{C}}\mathbbm{S}(\boldsymbol{u}_{\boldsymbol{x}})_{j}\log_{2}\mathbbm{S}(\boldsymbol{u}_{\boldsymbol{x}})_{j}$ \\
6 &  Class proportion~{\citep{LA}}  & $g_{\boldsymbol{x}}^{6} = \frac{\mathcal{N}_{{y}_{\boldsymbol{x}}}}{\mathcal{N}}$ \\
7 & The average loss of each category  & $g_{\boldsymbol{x}}^{7} = \bar{\ell}_{{y}_{\boldsymbol{x}}}$ \\
8 & The norm of the classifier weights~{\citep{classifier}}  & $g_{\boldsymbol{x}}^{8} = ||\boldsymbol{W}^{\mathcal{C}}_{{y_{\boldsymbol{x}}}}||_{2}^{2}$ \\
9 &  Relative loss  & $g_{\boldsymbol{x}}^{9} = \ell_{\boldsymbol{x}}-\bar{\ell}_{{y}_{\boldsymbol{x}}}$ \\
10 &  Relative margin  & $g_{\boldsymbol{x}}^{10} = g_{\boldsymbol{x}}^{2}-\bar{\gamma}_{{y}_{\boldsymbol{x}}}$ \\

\bottomrule
\end{tabular}
} 
\caption{{Extracted ten training characteristics of samples.}}
\label{chara} 
\end{table}

The calculation formulas for these ten quantities are detailed in Table~\ref{chara}, which are further explained below. First and foremost, we consider four factors used to measure the learning difficulty of individual samples. Sample loss ($\ell_{\boldsymbol{x}}$) and margin are always utilized for this purpose. Margin generally measures the distance from a sample to the classification boundary. A larger loss and a smaller margin indicate a greater learning difficulty. In the calculation formula for margin, $\mathbbm{S}(\cdot)$ refers to the Softmax function, and $\boldsymbol{u}_{\boldsymbol{x}}$ denotes the logit vector for sample $\boldsymbol{x}$. Moreover, the norm of the loss gradient for the logit vector $\boldsymbol{u}_{\boldsymbol{x}}$ is another commonly used training characteristic to measure learning difficulty. In its calculation, $\boldsymbol{y}_{\boldsymbol{x}}$ is the one-hot label vector of sample $\boldsymbol{x}$. Furthermore, the learning difficulty of samples can also be reflected by the cosine value ($\cos \theta_{\boldsymbol{x},y_{\boldsymbol{x}}}$) of the angle between the feature and the weight vector, which is derived as follows:
\begin{equation}
\begin{aligned}
\mathcal{F}(\boldsymbol{x})_{y_{\boldsymbol{x}}} &:={\boldsymbol{W}_{y_{\boldsymbol{x}}}^{{\mathcal{C}}}}^{T} \boldsymbol{x}+b_{y_{\boldsymbol{x}}}  \\
&:=\left\|\boldsymbol{W}_{y_{\boldsymbol{x}}}^{{\mathcal{C}}}\right\|\left\|\boldsymbol{x}\right\| \cos \theta_{\boldsymbol{x},y_{\boldsymbol{x}}} \ \text{by omitting bias term}\\
&:=\left\|\boldsymbol{x}\right\| \cos \theta_{\boldsymbol{x},y_{\boldsymbol{x}}} \ \text{by weight normalization}\\
&:=\cos \theta_{\boldsymbol{x},y_{\boldsymbol{x}}}, \ \text{by feature normalization}
\end{aligned}
\end{equation}
where $\boldsymbol{W}^{{\mathcal{C}}}_{y_{\boldsymbol{x}}}$ represents the classifier weights, $\boldsymbol{W}^{{\mathcal{C}}}$, of class ${y_{\boldsymbol{x}}}$.
Subsequently, the uncertainty of training samples is also taken into account, calculated by the information entropy of the Softmax output, where $\mathcal{C}$ refers to the number of classes. Additionally, three factors are considered to measure the learning difficulty of each class. The class proportion is the most commonly used factor to address imbalanced classification. In its calculation, $\mathcal{N}_{{y}_{\boldsymbol{x}}}$ and $\mathcal{N}$ represent the numbers of samples in class ${y}_{\boldsymbol{x}}$ and in the entire training set, respectively. The average loss of each category is another class-level factor, indicating the average learning difficulty of samples in each class. In its calculation, $\bar{\ell}_{{y}_{\boldsymbol{x}}}$ denotes the average loss of samples in class ${y}_{\boldsymbol{x}}$. Moreover, the norm of the classifier weights is a technique proven useful in handling long-tailed data. Finally, two relative factors are considered to reflect the learning difficulty of each sample relative to the average learning difficulty of its class, including relative loss and relative margin, where $\overline{\gamma}_{y_{\boldsymbol{x}}}$ denotes the average margin of samples in class $y_{\boldsymbol{x}}$.

Consequently, the ten characteristics mentioned above will be extracted from the classifier and utilized as inputs for the perturbation network to generate logit perturbations for samples.

\begin{algorithm}[t]
    \caption{Training algorithm of CLP}
    \label{alg1}
    \textbf{Input}: Training data $\mathcal{D}^{train}$, augmented metadata $\mathcal{D}^{meta}$, batch sizes $n$ and $m$, step sizes $\eta_{1}$ and $\eta_{2}$, maximum iterations $\mathcal{T}$. \\
       \textbf{Output}: Learned $\boldsymbol{W}$ and $\boldsymbol{\Omega}$.
    \begin{algorithmic}[1]

\STATE {Initialize $\boldsymbol{W}^{1}$ and $\boldsymbol{\Omega}^{1}$;}
\FOR {$t = 1$ to $\mathcal{T}$}
    \STATE{Sample $n$ and $m$ samples from $\mathcal{D}^{train}$ and $\mathcal{D}^{meta}$;}
    \STATE{Formulate $\hat{\boldsymbol{W}}^{t}(\boldsymbol{\Omega})$ by Eq.~(\ref{meta1});}
    \STATE{Extracted the training characteristics $\boldsymbol{g}_{\boldsymbol{x}}$ of samples from the classifier;}
    \STATE{Update $\boldsymbol{\Omega}^{t+1}$ by Eq.~(\ref{meta2});}
    \STATE{Update $\boldsymbol{W}^{t+1}$ by Eq.~(\ref{meta3}) with the logit perturbation terms generated by the perturbation network;}
    \ENDFOR
    \end{algorithmic}
\end{algorithm}

\subsubsection{{Meta-Learning-Based Training Algorithm}}
Given a set of augmented metadata $\{(\boldsymbol{x}_{i}^{{meta}},y_{i}^{{meta}})\}_{i=1}^{\mathcal{M}}$, we introduce a meta-learning objective that incorporates both a classification loss and a saliency regularization term. The saliency term~\citep{saldefine} aims to weaken the associations between non-causal features and their corresponding labels by the model. Its computation is demonstrated below:
\begin{equation}
\mathcal{R}_{i}^{{Sal}}= \lambda\sum_{j=1}^\mathcal{G}\left(\frac{\partial \hat{\boldsymbol{y}}_{y_i}}{\partial \boldsymbol{x}_{i,j}}\right)^2 \times\left(1-\boldsymbol{r}_{i,{j}}\right) / \sum_{j=1}^\mathcal{G}\left(1-\boldsymbol{r}_{i,{j}}\right),
\end{equation}
where $\mathcal{G}$ refers to the number of pixels in an image and $\lambda$ is a hyperparameter. Besides, $\hat{y}_{y_i}$ refers to the predicted probability for class $y_i$. $\boldsymbol{r}_{i,j}$ indicates whether the $j$th pixel in $\boldsymbol{x}_{i}$ belongs to the causal area. Then, our learning problem can be formulated as the following bi-level optimization problem:
\begin{equation}
\begin{aligned}
\mathop {\min }\limits_{\boldsymbol{\Omega}}  \sum\limits_{j = 1}^\mathcal{M} &{\left[\hat{\ell}\left({\mathbbm{S}\rm{(}}\boldsymbol{u}_j^{{meta}}({\boldsymbol{W}^*}) + {\boldsymbol{\delta}^{{meta}}_j}{\rm{(}}\boldsymbol{\Omega}{\rm{))}}, y_j^{{{meta}}}\right)+\mathcal{R}_{j}^{{Sal}}\right]},
\\
\text{s.t.},{\boldsymbol{W}^*}(\boldsymbol{\Omega}) &= \arg \mathop {\min}\limits_{\boldsymbol{W}}  \sum\limits_{i=1}^{\mathcal{N}} {\left[\ell\left({\mathbbm{S}(}{\boldsymbol{u}_i}(\boldsymbol{W}) + {\boldsymbol{\delta}_i}{(}\boldsymbol{\Omega}{))}, {y_i}\right)+\mathcal{R}_{i}^{{Sal}}\right]},
\end{aligned}
\label{question}
\end{equation}
where $\boldsymbol{W}$ and $\boldsymbol{\Omega}$ refer to the parameters in the classifier and the perturbation network. {$\hat{\ell}$ and $\ell$ represent the training losses on the meta data and the training data, respectively.} $\mathcal{M}$ and $\mathcal{N}$ denote the number of samples in the meta and training datasets. Besides, $\boldsymbol{u}_{i}$ and $\boldsymbol{\delta}_{i}$ represent the logit vector and logit perturbation for sample $\boldsymbol{x}_{i}$. It is worth noting that even when the meta dataset is lacking, it can be effectively generated by extracting meaningful samples from the training data. As the augmented data contains counterfactuals, the corresponding loss function $\hat{\ell}$ is defined as $\ell^{{CF}}$ when a given sample is a counterfactual, and it dissolves into the standard CE loss otherwise. Eq.~(\ref{question}) is a challenging problem, and we apply an online meta-learning-based learning {algorithm} to alternatively update the parameters of the classifier and the perturbation network (i.e., $\boldsymbol{W}$ and $\boldsymbol{\Omega}$) during training,
as illustrated in Fig.~\ref{frame}.

First, we treat $\boldsymbol{\Omega}$ as the parameter to be updated and formulate the parameter of the updated classifier $\boldsymbol{W}$, which is a function of $\boldsymbol{\Omega}$. Stochastic gradient descent (SGD) is employed to optimize the training loss. In each iteration, a minibatch of training samples $\{(\boldsymbol{x}_{i},{y}_{i})\}_{i=1}^{n}$ is selected, where $n$ refers to the minibatch size. The update of $\boldsymbol{W}$ can be formulated as follows:
\begin{equation}
\begin{aligned}
\hat{\boldsymbol{W}}^{t}\leftarrow\boldsymbol{W}^{t}-\eta_{1} \frac{1}{n} \sum_{i=1}^{n} \nabla_{_{\boldsymbol{W}}}\{&\ell(\mathbbm{S}(\boldsymbol{u}_{i} (\boldsymbol{W}^{t})+\boldsymbol{\delta}_{i}(\boldsymbol{\Omega})),y_{i})+ \mathcal{R}_{i}^{{Sal}}\},
\end{aligned}
\label{meta1}
\end{equation}
where $\eta_{1}$ is the step size. 
The parameters of the perturbation network ($\boldsymbol{\Omega}$) can be updated on a minibatch of metadata after receiving feedback from the classifier network, as follows:
\begin{equation}
\begin{aligned}
\boldsymbol{\Omega}^{t+1}\leftarrow \boldsymbol{\Omega}^{t}-\eta_{2} \frac{1}{m} \sum_{i=1}^{m} \nabla_{\boldsymbol{\Omega}}\{&\hat{\ell}(\mathbbm{S}(\boldsymbol{u}_{i}^{{meta}}(\hat{\boldsymbol{W}}^{t})+\boldsymbol{\delta}_{i}^{{meta}}(\boldsymbol{\Omega}^{t})),y_{i}^{{meta}})+ \mathcal{R}_{i}^{{Sal}}\},
\end{aligned}
\label{meta2}
\end{equation}
where $m$ and $\eta_{2}$ are the minibatch size of metadata and the step size, respectively. The parameters of the classifier network are finally updated with the generated perturbations, by fixing the parameters of the perturbation network as $\boldsymbol{\Omega}^{t+1}$:
\begin{equation}
\begin{aligned}
\boldsymbol{W}^{t+1}\leftarrow\boldsymbol{W}^{t}-\eta_{1} \frac{1}{n} \sum\nolimits_{i=1}^{n}\nabla_{_{\boldsymbol{W}}}\{&\ell(\mathbbm{S}(\boldsymbol{u}_{i}(\boldsymbol{W}^{t})+\boldsymbol{\delta}_{i}(\boldsymbol{\Omega}^{t+1})),y_{i})+\mathcal{R}_{i}^{{Sal}}\}.
\end{aligned}
\label{meta3}
\end{equation}
The meta-learning-based {training} algorithm of our CLP framework is presented in Algorithm~1.

\section{Experimental Investigation}
Extensive experiments are conducted to evaluate the efficacy of the proposed CLP framework across four typical learning scenarios impacted by data biases, including LT learning, noisy label learning, subpopulation shift learning, and GLT learning.
Each experiment is repeated three times with different seeds. The hyperparameter $\lambda$ which controls the strength of the regularization term is selected from the set $\{0.2, 0.4, 0.6, 0.8, 1\}$. Throughout all experiments, the number of augmented samples is consistently set to $2\hat{\mathcal{M}}$, where $\hat{\mathcal{M}}$ denotes the size of the original metadata. In scenarios where both counterfactual and factual augmentations are utilized, the ratio of samples generated by the two manners is 1:1. Furthermore, to assess CLP's effectiveness in breaking spurious associations between background and labels, we conduct a comparative analysis with a variant of CLP, termed Meta Logit Perturbation (Meta-LP). Compared to CLP, Meta-LP omits causal augmentation for metadata.

\begin{table}[t]
\centering
\small

\resizebox{1\textwidth}{!}{
\begin{tabular}{l|c|c|c|c}
\toprule
Dataset & \multicolumn{2}{c|}{CIFAR10} & \multicolumn{2}{c}{CIFAR100} \\ \hline
Imbalance ratio                      & 100:1            & 10:1      & 100:1            & 10:1        \\ \hline\hline
 Class-Balanced CE loss     & 27.32\%          &13.10\%     & 61.23\%          & 42.43\%       \\
Class-Balanced Fine-Tuning& 28.66\%           & 16.83\%    & 58.50\%          & 42.43\%      \\
Meta-Weight-Net            & 26.43\%   & 12.45\%   & 58.39\%          & 41.09\%      \\
Focal loss                 & 29.62\%   & 13.34\%    & 61.59\%          & 44.22\%          \\
Class-Balanced Focal loss  & 25.43\%   & 12.52\%     & 60.40\%          & 42.01\%       \\
LDAM                      & 26.45\% & 12.68\%     & 59.40\%          & 42.71\%       \\
LDAM-DRW                  & 21.88\%   & 11.63\%     & 57.11\%          & 41.22\%        \\
ISDA             & 27.45\%   & 12.98\%     & 62.60\%          & 44.49\%    \\
LA                        & 22.33\%  & 11.07\%       & 56.11\%          & 41.66\%      \\
MetaSAug                       & \underline{19.46}\%      & 10.56\%      & \underline{53.13}\%      & 38.27\%    \\
ALA                       & 21.97\%      & 10.64\%    & 54.56\%          & 39.42\%      \\
LPL                        & 22.05\%      & 10.59\%     & 55.75\%          & 39.03\%      \\
{CSA}                        & {19.47\%}      & {\underline{10.20}\%}     & {53.39\%}          & {\underline{37.40}\%}      \\
\hline
 \cellcolor{gray!25}{Meta-LP} & \cellcolor{gray!25}{21.86\%} & \cellcolor{gray!25}{10.24\%} & \cellcolor{gray!25}{54.23\%} & \cellcolor{gray!25}{38.79\%}  \\ 
 \cellcolor{gray!25}{CLP}  &\cellcolor{gray!25}\textbf{18.03}\%          & \cellcolor{gray!25}\textbf{9.08}\%      & \cellcolor{gray!25}\textbf{51.66}\%          & \cellcolor{gray!25}\textbf{36.31}\%     \\ 
\bottomrule
\end{tabular}
}
\caption{Top-1 error rate on CIFAR-LT {using the} ResNet-32 model. The best and the second-best results are highlighted in bold and underlined, respectively.}
\label{table1}
\end{table}

\begin{table}
\begin{tabular}{l|c|c|c|c}
\toprule
Dataset & \multicolumn{2}{c|}{CIFAR10} & \multicolumn{2}{c}{CIFAR100} \\ \hline
Imbalance ratio                      & 100:1            & 10:1      & 100:1            & 10:1        \\ \hline\hline
 CF(Grey)                  & {19.25}\%& {10.05}\%  & {52.88\%} & {37.47\%}  \\
CF(Shuffle)               & 19.05\%          & 9.94\%      & 52.64\%          & 37.54\%     \\
CF(Tile)                 & 19.67\%          & 9.89\%     & 52.97\%          & 37.69\%     \\ 
CF(CAGAN)                 & 19.45\%          & 9.73\%      & 52.89\%          & 37.06\%     \\ 
F(Shuffle)                 & 18.43\%          & 9.45\%      & 53.03\%          & 37.24\%       \\ 
F(Random)                 & \underline{18.34}\%          & 9.78\%   & 52.31\%          & 37.13\%           \\ 
F(Mix-Rand)                 & 19.05\%          & 9.65\%       & 52.17\%          & 36.96\%       \\ 
F(FGSM $\epsilon=0.5$)                 & 18.79\%          & 9.42\%       & 52.58\%          & 36.87\%     \\ 
CF(Tail)+F(Shuffle)                 & \underline{18.34}\%          & \textbf{9.08}\%    & \underline{51.79}\%          & \underline{36.44}\%       \\ 
CF(CAGAN)+F(Random)                 & \textbf{18.03}\%          & \underline{9.25}\%      & \textbf{51.66}\%          & \textbf{36.31}\%     \\  \bottomrule
\end{tabular}
\caption{{Comparison of various data augmentation manners in our CLP framework on CIFAR-LT using the ResNet-32 model. CF and F denote the counterfactual and factual augmentations, respectively.}}
\label{table11}
\end{table}

\subsection{Experiments on Long-Tailed Classification}

The presence of long-tailed class distribution is a prevalent issue in real-world datasets. 
Two long-tailed benchmarks, including CIFAR-LT~\citep{classimb} and ImageNet-LT~\citep{imagenetLT}, are taken into account. Due to space constraints, only experiments on CIFAR-LT are provided here.
We discard some training samples to construct the long-tailed CIFAR datasets. 
The training and testing configurations employed by Shu et al. are followed. We train the ResNet-32~\citep{reshe} model with an initial learning rate of 0.1 and the standard SGD with a momentum of 0.9 and a weight decay of $5\times 10^{-4}$. The learning rate is decayed by 0.1 at the $120$th and $160$th epochs. 
{Ten images per category are randomly selected from the validation set to compile the metadata.} We compare our proposed CLP using different augmentation manners with various {advanced methods that are tailored for LT learning}, including Class-Balanced CE loss~\citep{classimb}, Class-Balanced Fine-Tuning, Meta-Weight-Net, Focal loss~\citep{Focal}, Class-Balanced Focal loss, Label-Distribution-Aware Margin Loss (LDAM), LDAM-DRW, Implicit Semantic Data Augmentation (ISDA)~\citep{ISDA}, Logit Adjustment (LA), MetaSAug~\citep{metasaug}, Adaptive Logit Adjustment (ALA), {Context Shift Augmentation (CSA)~\citep{CSA},} and Logit Perturbation Loss (LPL).

\begin{figure*}[t] 
\centering
\includegraphics[width=0.9\textwidth]{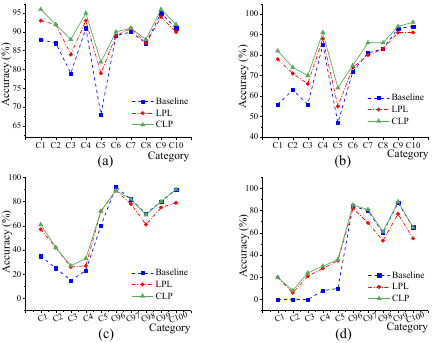}
\caption{Class-wise accuracy on CIFAR10-LT with imbalance ratios of 10:1 (a) and 100:1 (b). Class-wise accuracy on CIFAR100-LT with imbalance ratios of 10:1 (c) and 100:1 (d). From ``C1" to ``C10", the number of samples varies from small to large.}
\label{imb}
\end{figure*}

The Top-1 error rates of CLP using different augmentation techniques, Meta-LP, and other LT baselines on CIFAR10-LT and CIFAR100-LT are {presented} in Table~\ref{table1}. CLP exhibits a significant decrease in error rates on CIFAR10-LT, achieving reductions of $1.43\%$ and {$1.12\%$} for imbalance ratios of 100:1 and 10:1, respectively, compared to the optimal outcomes of baselines. Similarly, on CIFAR100-LT, CLP achieves lowered error rates of $1.47\%$ and {$1.09\%$} for imbalance ratios of 100:1 and 10:1, respectively. Fig.~\ref{imb} illustrates that compared to LPL, CLP not only improves the accuracy of tail categories but also enhances the accuracy of head categories. Additionally, {Meta-LP outperforms a range of compared baselines on imbalanced datasets (e.g., ALA and LPL).} This is attributed to the balanced metadata set, which guides the perturbation network to generate sample perturbations that are beneficial for balanced datasets. Nevertheless, CLP outperforms Meta-LP, indicating the presence of spurious associations between backgrounds and labels in classifiers trained on imbalanced CIFAR data, which adversely affects model performance. Crucially, CLP effectively assists the classifier in breaking these spurious associations, contributing to its superior performance. 

{The comparison of CLP with different data augmentation methods is presented in Table~\ref{table11}.} Notably, when both counterfactual and factual augmentations are applied, the proposed CLP achieves the lowest error rates. Furthermore, even when employing a single augmentation strategy, CLP demonstrates competitive performance. Specifically, using only factual augmentation generally yields better results than using counterfactual augmentation alone, owing to the enlargement of images with consistent foregrounds, enabling the classifier to acquire more generalizable features. {Moreover, our experimental results indicate that once the meta dataset is compiled and augmented, our method's training time is comparable to that of previous meta-learning methods, such as Meta-Weight-Net and MetaSAug, as compared in Table~\ref{time}. However, the proposed CLP framework achieves superior performance, thereby ensuring a better balance between computational efficiency and performance.}
\begin{table}[t]
\centering
\begin{tabular}{l|cc}
\toprule
                & CIFAR10 & CIFAR100 \\ \midrule
Meta-Weight-Net &   189s      &   192s       \\
MetaSAug        &   198s      &     197s     \\
 \rowcolor{gray!25} CLP             &    195s     &   194s      \\ \bottomrule
\end{tabular}
\caption{{Comparison of training time per epoch between our approach and previous meta-learning-based baselines.}}
\label{time}
\end{table}



\begin{table}[t]
\centering
\small

\resizebox{1\textwidth}{!}{
\begin{tabular}{l|cc|cc|cc|cc}
\toprule
Noise type & \multicolumn{4}{c|}{Flip} &  \multicolumn{4}{c}{Uniform} \\ \hline
                    Dataset   & \multicolumn{2}{c|}{CIFAR10}                         & \multicolumn{2}{c|}{CIFAR100}  & \multicolumn{2}{c|}{CIFAR10}                         & \multicolumn{2}{c}{CIFAR100}                      \\ \hline
Noise ratio            & \multicolumn{1}{c|}{20\%}           & 40\%           & \multicolumn{1}{c|}{20\%}           & 40\%   & \multicolumn{1}{c|}{40\%}           & 60\%           & \multicolumn{1}{c|}{40\%}           & 60\%        \\ \hline\hline
CE loss & \multicolumn{1}{l|}{76.83\%}     & 70.77\%     & \multicolumn{1}{l|}{50.86\%}     & 43.01\% & \multicolumn{1}{l|}{68.07\%}     & 53.12\%     & \multicolumn{1}{l|}{51.11\%}     & 30.92\%    \\
Focal loss             & \multicolumn{1}{l|}{86.45\%}     & 80.45\%     & \multicolumn{1}{l|}{61.87\%}     & 54.13\%  & \multicolumn{1}{l|}{75.96\%}     &   51.87\%  & \multicolumn{1}{l|}{51.19\%}     & 27.70\%   \\
Co-Teaching           & \multicolumn{1}{l|}{82.83\%}     & 75.41\%     & \multicolumn{1}{l|}{54.13\%}     & 44.85\%   & \multicolumn{1}{l|}{74.81\%}     & 73.06\%     & \multicolumn{1}{l|}{46.20\%}     & 35.67\%  \\
APL                    & \multicolumn{1}{l|}{87.23\%}     & 80.08\%     & \multicolumn{1}{l|}{59.37\%}     & 52.98\%  & \multicolumn{1}{l|}{86.49\%}     & 79.22\%     & \multicolumn{1}{l|}{57.84\%}     & 49.13\%  \\
MentorNet              & \multicolumn{1}{l|}{86.36\%}     & 81.76\%    & \multicolumn{1}{l|}{61.97\%}     & 52.66\%   & \multicolumn{1}{l|}{87.33\%}     & 82.80\%    & \multicolumn{1}{l|}{61.39\%}     & 36.87\%  \\
L2RW                  & \multicolumn{1}{l|}{87.86\%}     & 85.66\%     & \multicolumn{1}{l|}{57.47\%}     & 50.98\%     & \multicolumn{1}{l|}{86.92\%}     & 82.24\%     & \multicolumn{1}{l|}{60.79\%}     & 48.15\% \\
GLC                   & \multicolumn{1}{l|}{89.58\%}     & 88.92\%     & \multicolumn{1}{l|}{63.07\%}     & 59.22\%   & \multicolumn{1}{l|}{88.28\%}     & 83.49\%     & \multicolumn{1}{l|}{61.31\%}     & 50.81\%     \\
Meta-Weight-Net       & \multicolumn{1}{l|}{90.33\%}     & 87.54\%     & \multicolumn{1}{l|}{64.22\%}     & 58.64\%    & \multicolumn{1}{l|}{89.27\%}     & 84.07\%     & \multicolumn{1}{l|}{67.73\%}     & 58.75\%  \\ 
$\mathcal{L}_{DMI}$        & \multicolumn{1}{l|}{86.70\%}     & 84.00\%     & \multicolumn{1}{l|}{62.26\%}     & 57.23\%    & \multicolumn{1}{l|}{85.90\%}     & 79.60\%     & \multicolumn{1}{l|}{63.16\%}     & 55.37\%  \\ 
JoCoR       & \multicolumn{1}{l|}{{90.78\%}}     & 83.67\%     & \multicolumn{1}{l|}{65.21\%}     & 45.44\%   & \multicolumn{1}{l|}{89.15\%}     & 64.54\%     & \multicolumn{1}{l|}{65.45\%}     & 44.43\%   \\ 
ISDA     & \multicolumn{1}{l|}{88.90\%}     & 86.14\%     & \multicolumn{1}{l|}{64.36\%}     & 59.48\%   & \multicolumn{1}{l|}{88.11\%}     & 83.12\%     & \multicolumn{1}{l|}{65.15\%}     & 58.19\%   \\ 
MetaSAug    & \multicolumn{1}{l|}{90.42\%}     & 87.73\%  & \multicolumn{1}{l|}{66.47\%}     & 61.43\%   & \multicolumn{1}{l|}{89.32\%}     & 84.65\% & \multicolumn{1}{l|}{66.50\%}     & 59.84\%   \\ 
{MFRW-MES}     & \multicolumn{1}{l|}{{\underline{91.05}\%}}     & {\underline{89.42}\%}  & \multicolumn{1}{l|}{{65.27\%}}     & {61.85\%}   & \multicolumn{1}{l|}{{\underline{89.68}\%}}     & {\underline{85.03}\%} & \multicolumn{1}{l|}{{68.08\%}}     & {59.69\%}   \\ 
\hline
\rowcolor{gray!25} Meta-LP &\multicolumn{1}{l|}{\underline{91.05}\%} & 89.34\% & \multicolumn{1}{l|}{\underline{66.85}\%} & \underline{61.86}\% & \multicolumn{1}{l|}{89.58\%} & 84.87\% & \multicolumn{1}{l|}{\underline{68.09}\%} & \underline{60.26}\% \\ 
\rowcolor{gray!25} CLP &\multicolumn{1}{l|}{\textbf{92.34}\%} & \textbf{90.36}\% & \multicolumn{1}{l|}{\textbf{67.98}\%} & \textbf{62.93}\% & \multicolumn{1}{l|}{\textbf{90.66}\%} & \textbf{85.77}\% & \multicolumn{1}{l|}{\textbf{69.01}\%} & \textbf{61.55}\% \\ 
\bottomrule
\end{tabular}
}
\caption{Top-1 accuracy on CIFAR data with flip and uniform noise.}
\label{table4}
\end{table}

\subsection{Experiments on Noisy Label Classification}

\begin{figure}[t] 
\centering
\includegraphics[width=0.95\textwidth]{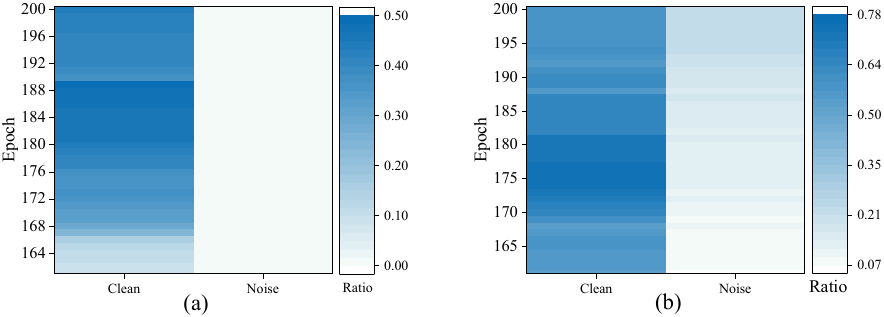}
\caption{Ratio of samples in CIFAR10 where perturbations from noise and clean samples result in an increase in losses during training. (a) and (b) respectively illustrate the scenarios with 20\% and 40\% flip noise. {The proportion of clean samples with increased losses is significantly higher than that of noisy samples, thereby reducing the adverse impact of noisy samples on model training.}}
\label{noisefenxi}
\end{figure}

To evaluate the effectiveness of CLP in noisy label learning scenarios, we adopt two settings of corrupted labels: uniform and pair-flipped noisy labels. Uniform noise refers to the label of each
sample being independently changed to a random class with a fixed probability. Flip noise refers to the label of each sample independently flipped to the two most similar classes with a total
probability. The CIFAR10 and CIFAR100~\citep{cifar} datasets are utilized for this purpose, due to their widespread use in noisy label evaluation. We follow Shu et al. to construct metadata, wherein 1,000 images with clean labels from the validation set are selected as metadata. The ratios for flip noise are set to 20\% and 40\%, while those for uniform noise are set to 40\% and 60\%. For the uniform and pair-flip noise, Wide ResNet-28-10 (WRN-28-10)~\citep{wideres} and ResNet-32 are used as classifiers, respectively. The initial learning rate and batch size are set to 0.1 and 128, respectively. For ResNet, standard SGD with a momentum of 0.9 and a weight decay of $1\times 10^{-4}$ is utilized. For Wide ResNet, standard SGD with a momentum
of 0.9 and a weight decay of $5\times 10^{-4}$ is utilized.
We compare the CLP framework using different augmentation manners with several advanced methods designed for noisy learning, including the baseline method, which trains the backbone network with CE loss; robust learning methods such as Focal loss, Co-Teaching~\citep{coteaching}, Active Passive Loss (APL)~\citep{APL}, ISDA, Joint Training Method with Co-Regularization (JoCoR)~\citep{Jocor}, and Information-Theoretic Loss ($\mathcal{L}_{DMI}$)~\citep{LDMI}; and meta-learning-based approaches such as MentorNet~\citep{LuJiang41}, Learning to Reweight (L2RW)~\citep{L2RW}, Gold Loss Correction (GLC)~\citep{GLC}, Meta-Weight-Net, {Meta Feature Re-Weighting with Meta Equalization Softmax (MFRW-MES)~\citep{mfrw}}, and MetaSAug.

Table~\ref{table4} presents the Top-1 accuracy achieved on CIFAR10 and CIFAR100 datasets with flip and uniform noise, respectively, demonstrating the SOTA performance of CLP across all cases. In the context of flip noise, our method outperforms the best performance among all compared methods by {$1.29\%$} and {$0.94\%$}, respectively, when the noise ratio is set at $20\%$ and $40\%$ on CIFAR10. Similarly, on noisy CIFAR100, CLP boosts accuracies by $1.51\%$ and {$1.08\%$}, respectively. Concerning uniform noise, our approach yields accuracy surpassing the optimal accuracy achieved by other methods by {$0.98\%$} and {$0.74\%$} on noisy CIFAR10 when the noise ratio is $40\%$ and $60\%$, respectively. Furthermore, on noisy CIFAR100, CLP exhibits an increase in accuracies by {$0.93\%$} and $1.71\%$, respectively. 

{As shown in Fig.~\ref{noisefenxi}, the proportion of clean samples experiencing increased losses due to logit perturbations is significantly higher than that of noisy samples. This indicates that the perturbations generated by the CLP framework can effectively reduce the negative impact of noisy samples, thereby enhancing model performance.} Additionally, while Meta-LP also showcases promising results on noisy CIFAR data as it relies on a clean meta dataset, CLP surpasses Meta-LP in performance. This indicates the existence of spurious correlations between image backgrounds and class labels in models trained on noisy CIFAR data. Consequently, CLP demonstrates its efficacy in breaking spurious correlations and tackling label noise.

\subsection{Experiments on Subpopulation Shift Classification}

\begin{table}[t]
\centering
\begin{tabular}{l|l|cc}
\toprule
& Method & Avg. $\uparrow$ & Worst $\uparrow$\\ \hline\hline
\multirow{9}{*}{Others} & UW       & \textbf{95.1}\%          & 88.0\%               \\
& IRM       & 87.5\%          & 75.6\%         \\
& IB-IRM    & 88.5\%           & 76.5\%        \\
& V-REx     & 88.0\%           & 73.6\%        \\
& CORAL     & 90.3\%           & 79.8\%               \\
& GroupDRO  & 91.8\%           & \underline{90.6}\%            \\
& DomainMix & 76.4\%           & 53.0\%         \\
& LISA      & 91.8\%           & 89.2\%           \\ 
& {\textsc{CnC}}      & {90.8\%}           & {88.5\%}           \\ 
& {ACE}      & {90.8\%}           & {75.1\%}           \\
& {PDE}      & {92.0\%}           & {90.5\%}           \\ \hline
\multirow{2}{*}{Ours} & \cellcolor{gray!25}Meta-LP & \cellcolor{gray!25}91.3\% & \cellcolor{gray!25}90.1\% \\ 
&  \cellcolor{gray!25}CLP & \cellcolor{gray!25}\underline{93.3}\% & \cellcolor{gray!25}\textbf{91.8}\% \\
\bottomrule
\end{tabular}
\caption{The average Top-1 and the worst-group accuracy on Waterbirds dataset using the ResNet-50 model.}
\label{table6}
\end{table}

To evaluate the capability of CLP in breaking spurious correlations caused by subpopulation shifts, we conduct experiments on the Waterbirds~\citep{r54} dataset, which aims to classify birds as waterbird or landbird using the image background as the spurious attribute. Waterbirds is a synthetic dataset that combines bird images from the CUB dataset~\citep{r86} with backgrounds sampled from the Places dataset~\citep{r87}. Two groups including (``land" background, ``waterbird") and (``water" background, ``landbird") are regarded as minority groups. There are 4,795 training samples of which 56 samples are ``waterbirds on land" and 184 samples are ``landbirds on water". The remaining training samples include 3,498 samples from ``landbirds on land", and 1,057 samples from ``waterbirds on water". The hyperparameter settings follow those of Yao et al.~\citep{r57}. The pre-trained ResNet-50~\citep{reshe} is utilized as the backbone network with an initial learning rate of $1\!\times\!10^{-3}$ and a batch size of 16. We use the SGD optimizer with a weight decay of $1\!\times\!10^{-4}$.

Robust methods, including Invariant Risk Minimization (IRM)~\citep{r58}, Information Bottleneck Invariant risk minimization (IB-IRM)~\citep{r59}, Variance Risk Extrapolation (V-REx)~\citep{r60}, 
Correlation Alignment (CORAL)~\citep{DORAL}, 
Group Distributionally
Robust Optimization (GroupDRO)~\citep{r54}, Domain Mixup (DomainMix)~\citep{r65}, Learn Invariant
Predictors via Selective Augmentation (LISA)~\citep{r57}, {Correct-N-Contrast (\textsc{CnC})~\citep{CNC}, Augmentation by Counterfactual Explanation (ACE)~\citep{ACE}, and Progressive Data Expansion (PDE)~\citep{PDE},} are compared. We also include Upweighting (UW) for comparison since it is well-suited for subpopulation shifts. Ten samples are randomly sampled from each group of training data to compile metadata. 
In order to conduct a more comprehensive comparison, we report both the average and the worst-group accuracy to compare the performance.

Table~\ref{table6} presents the average and worst-group accuracy of all methods. Among various advanced invariant feature learning approaches, CLP demonstrates the highest worst-group accuracy, indicating its superiority in addressing subpopulation shifts. Additionally, the average accuracy of CLP is found to be competitively high. While Meta-LP also exhibits competitive performance, it falls short compared to CLP. The reason lies in the effectiveness of CLP in alleviating spurious associations between backgrounds and labels, which leads to improved robustness against subpopulation shifts.

\begin{table*}[t]
\centering
\small
\resizebox{1\textwidth}{!}{
\begin{tabular}{l|cc|cc|cc}
\toprule
Protocol     & \multicolumn{2}{c|}{CLT}  & \multicolumn{2}{c|}{GLT} & \multicolumn{2}{c}{ALT} \\ \hline
Method   & Acc. $\uparrow$        & Prec. $\uparrow$      & Acc. $\uparrow$        & Prec. $\uparrow$      & Acc. $\uparrow$        & Prec. $\uparrow$      \\ \hline\hline
 CE loss    &42.52\%  &47.92\%   & 34.75\%       & 40.65\%      & 41.73\%       & 41.74\%      \\
cRT       &45.92\% &45.34\%     & 37.57\%       & 37.51\%     & 41.59\%       & 41.43\%      \\
LWS        &46.43\% &45.90\%    & 37.94\%       & 38.01\%      & 41.70\%       & 41.71\%      \\
De-confound-TDE& 45.70\% &44.48\% &37.56\%       & 37.00\%      & 41.40\%       & 42.36\%      \\
BLSoftmax &  45.79\% & 46.27\%   & 37.09\%       & 38.08\%      & 41.32\%       & 41.37\%      \\
LA&   46.53\% & 45.56\%  & 37.80\%       & 37.56\%      & 41.73\%          & 41.74\%         \\
LDAM  &      {46.74}\% & 46.86\%   & {38.54}\%       & 39.08\%      & 42.66\%       & 41.80\%      \\
IFL        & 45.97\% & 52.06\%    & 37.96\%       & 44.47\%      & 45.89\%       & {46.42}\%      \\
RandAug     &46.40\% & {52.13}\%      & 38.24\% & {44.74}\%   & {46.29}\% &46.32\%   \\
ISDA    & {47.66}\% & 51.98\%& {39.44}\%   &  44.26\%  & {47.62}\% & {47.46}\% \\
RISDA     & {49.31}\% & 50.64\%& {38.45}\%   &  42.77\%  & {47.33}\% & {46.33}\% \\
MetaSAug     & 50.53\% &     55.21\% &  {41.27}\%         &   {47.38}\%      &    {49.12}\%         &   {48.56}\%          \\ 

{CSA}     & {{46.49}\%} & {50.77\%} & {{37.22}\%}   &  {42.01\%}  & {{43.03}\%} & {{44.05}\%} \\
\hline
 \cellcolor{gray!25}Meta-LP & \cellcolor{gray!25}\underline{50.87}\% & \cellcolor{gray!25}\underline{55.64}\% & \cellcolor{gray!25}\underline{41.56}\% & \cellcolor{gray!25}\underline{47.66}\% & \cellcolor{gray!25}\underline{49.46}\% & \cellcolor{gray!25}\underline{48.85}\% \\ 
\cellcolor{gray!25}CLP &  \cellcolor{gray!25}\textbf{52.01}\% & \cellcolor{gray!25}\textbf{56.72}\% & \cellcolor{gray!25}\textbf{43.03}\% & \cellcolor{gray!25}\textbf{48.97}\% & \cellcolor{gray!25}\textbf{51.12}\% & \cellcolor{gray!25}\textbf{50.38}\% \\

\bottomrule
\end{tabular}
}
\caption{Top-1 accuracy and precision of the CLT, GLT, and ALT protocols on the ImageNet-GLT benchmark using the ResNeXt-50 model.}
\label{table7}
\end{table*}

\begin{table*}[t]
\centering
\resizebox{1\textwidth}{!}{
\begin{tabular}{l|cc|cc|cc}
\toprule
 Protocol     & \multicolumn{2}{c|}{CLT}  & \multicolumn{2}{c|}{GLT} & \multicolumn{2}{c}{ALT} \\ \hline
Method   & Acc. $\uparrow$        & Prec. $\uparrow$      & Acc. $\uparrow$        & Prec. $\uparrow$      & Acc. $\uparrow$        & Prec. $\uparrow$      \\ \hline\hline
 CE loss    &72.34\% & 76.61\% & 63.79\% & 70.52\% & 50.17\% & 50.94\%      \\
cRT       &73.64\% & 75.84\% &  64.69\% & 68.33\% & 49.97\% & 50.37\%      \\
LWS        &72.60\% & 75.66\% & 63.60\% & 68.81\% & 50.14\% & 50.61\%      \\
De-confound-TDE & 73.79\% & 74.90\% & 66.07\% & 68.20\% &  50.76\% & 51.68\%      \\
BLSoftmax &  72.64\% & 75.25\% &  64.07\% & 68.59\% & 49.72\% & 50.65\%      \\
LA&   75.50\% & 76.88\% & 66.17\% & 68.35\% & 50.17\% & 50.94\%         \\
LDAM  &      75.57\% &  77.70\% & 67.26\% & 70.70\% & {55.52}\% & {56.21}\%      \\
IFL        & 74.31\% & 78.90\% & 65.31\% & 72.24\% & 52.86\% & 53.49\%      \\
RandAug     &{76.81}\% & {79.88}\% & {67.71}\% & {72.73}\% & 53.69\% & {54.71}\%   \\
ISDA   &{77.32}\%  & 79.23\% & 67.57\% & {72.89}\% & {54.43}\% &  54.62\%  \\
RISDA   &{76.34}\%  & 79.27\% & 66.85\% & {72.66}\% & {54.58}\% &  53.98\%  \\
MetaSAug     & 77.50\% &     79.43\% &  {67.07}\%         &   {72.68}\%      &    {54.66}\%         &   {54.73}\%          \\ 
{CSA}     & {74.25\%} &     {78.56\%} &  {{64.78}\%}         &   {{69.10}\%}      &    {{52.34}\%}         &   {{52.11}\%}          \\ 
\hline
  \rowcolor{gray!25}Meta-LP & \underline{77.71}\% & \underline{80.23}\% & \underline{67.92}\% & \underline{73.04}\% & \underline{55.76}\% & \underline{56.44}\% \\ 
   \rowcolor{gray!25}CLP & \textbf{78.83}\% & \textbf{80.92}\% & \textbf{68.87}\% & \textbf{74.05}\% & \textbf{56.89}\% & \textbf{57.78}\% \\ 
\bottomrule
\end{tabular}}
\caption{Top-1 accuracy and precision of the CLT, GLT, and ALT protocols on the MSCOCO-GLT benchmark using the ResNeXt-50 model.}
\label{table8}
\end{table*}

\subsection{Experiments on Generalized Long-{T}ail Classification}

\begin{figure}[t] 
\centering
\includegraphics[width=1\textwidth]{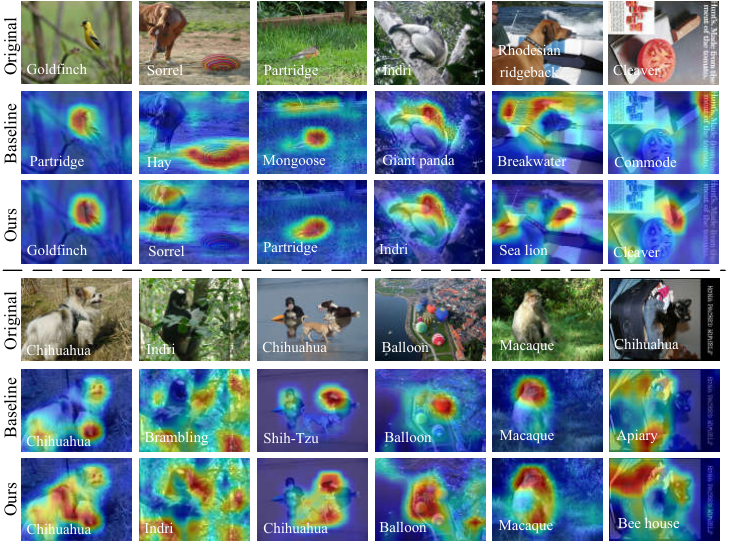}
\caption{Visualization of the regions that the model used for
making predictions. The colors blue and red represent regions that are indecisive and highly discriminative, respectively. The white text indicates both the true and predicted labels for each image.}
\label{visua}
\end{figure}

{Unlike traditional LT learning, which focuses solely on the imbalanced distribution of sample numbers across categories, GLT learning considers both the inter-category imbalance and the intra-category imbalance of attribute distributions.} In deep learning models, attribute imbalance is a more prevalent phenomenon compared to class imbalance, leading to two issues. First, due to the high rarity of certain attributes, the prediction performance of these samples is often poor. Second, attribute imbalance results in spurious associations between class labels and non-causal attributes. The efficacy of our approach in addressing attribute imbalance is demonstrated through the comparison of three protocols - CLT, GLT, and ALT, for the ImageNet-GLT and MSCOCO-GLT benchmarks. These protocols involve changing the class distribution, attribute distribution, and both class and attribute distributions from training to testing, respectively. 

The configurations for training and testing across the three protocols for the two benchmarks align with the specifications provided by Tang et al.~\citep{r67}. We employ ResNeXt-50~\citep{r37} as the backbone network for all methods, in calculating both the Top-1 accuracy and precision. The models undergo training at a batch size of 256, with an initial learning rate of 0.1, using the SGD optimizer with a weight decay of $5\times 10^{-4}$ and momentum of 0.9. Samples in metadata are extracted from each class in a balanced validation set assembled by Liu et al.~\citep{imagenetLT}. To collect metadata with attribute-wise balance, the images of each class are clustered into six groups using KMeans, and ten images per group and class are sampled to construct metadata. As for the compared methods, we study 
two-stage re-sampling
methods, including Classifier Re-Training (cRT)~\citep{r68} and Learnable Weight Scaling (LWS)~\citep{r68},
posthoc distribution adjustment methods including De-confound-TDE~\citep{r45} and LA, multi-branch models with diverse sampling
strategies like invariant feature learning methods like Invariant Feature Learning (IFL)~\citep{r67}, and
reweighting loss functions like Balanced Softmax (BLSoftmax)~\citep{r70}, {CSA},
and LDAM.
We also compare several data augmentation methods, including Random Augmentation (RandAug)~\citep{r71}, ISDA, RISDA, and MetaSAug. 

\begin{figure}[t] 
\centering
\includegraphics[width=0.95\textwidth]{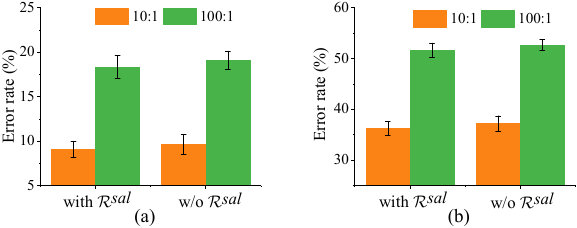}
\caption{Ablation studies for the saliency regularization term on CIFAR10 (a) and CIFAR100 (b) datasets with imbalance ratios of 10:1 and 100:1. Better results can be attained with the saliency regularization term.}
\label{salaba}
\end{figure}

\begin{figure}[t] 
\centering
\includegraphics[width=0.85\textwidth]{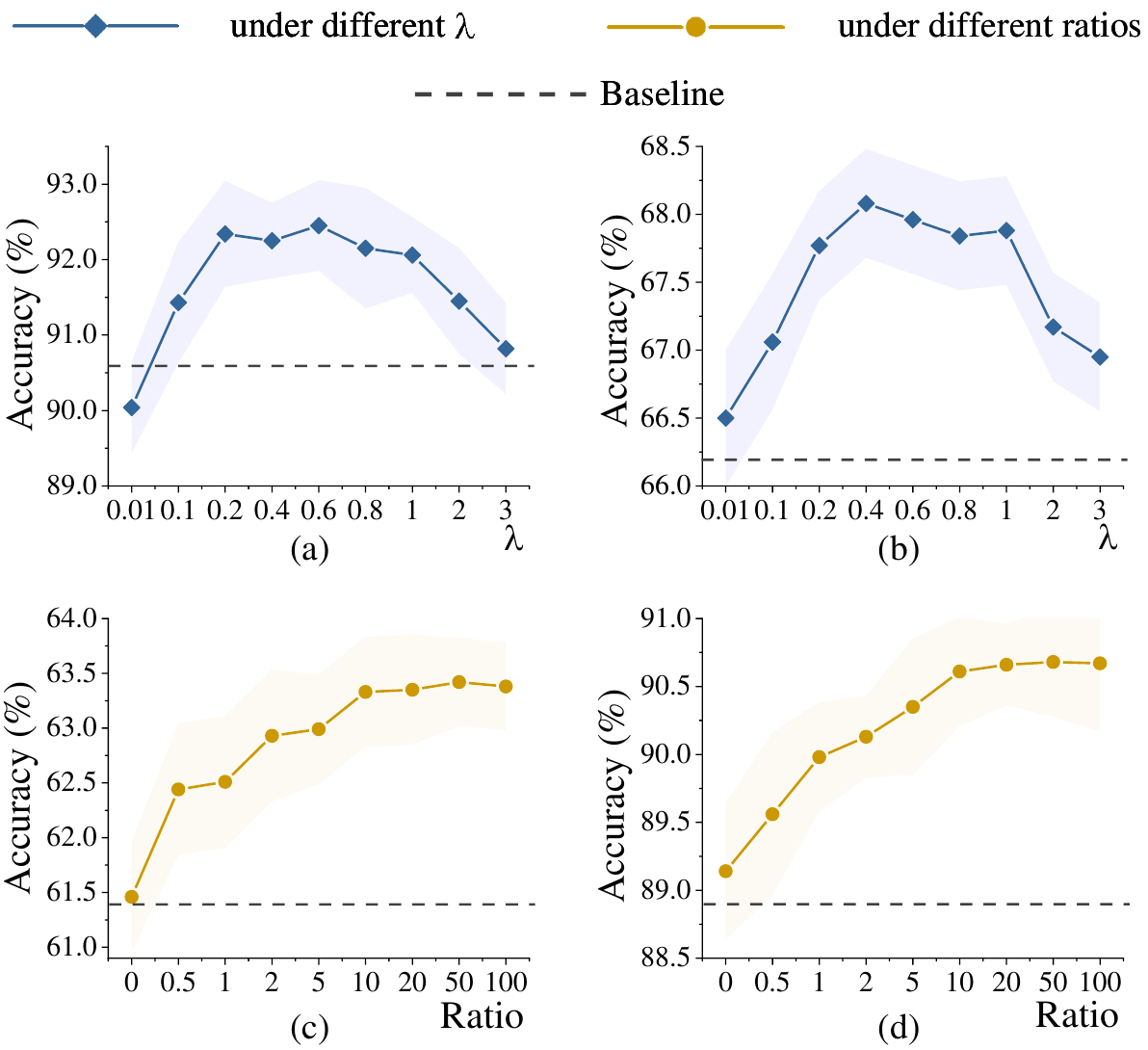}
\caption{Sensitivity tests of $\lambda$ on CIFAR10 (a) and CIFRA100 (b) datasets with 20\% flip noise. 
Sensitivity test of the number of augmented samples on CIFAR10 (c) and CIFAR100 (d) datasets with 40\% flip label noise. The horizontal axis represents the ratio of the number of augmented samples to that of the original meta samples. 
}
\label{sal2}
\end{figure}

Tables~\ref{table7} and~\ref{table8} present the results obtained on the three protocols of the ImageNet-GLT and MSCOCO-GLT benchmarks, respectively.
It is evident that all methods experience a decline in performance when transitioning from the CLT to the GLT protocol, signifying the greater challenge posed by addressing attribute-wise imbalance compared to class-wise imbalance. Remarkably, CLP enhances performance across all protocols, encompassing CLT, GLT, and ALT, which effectively addresses spurious correlations arising from imbalanced class and attribute distributions. In contrast, traditional methods that rely on rebalancing strategies to tackle long-tailed datasets prove ineffective in improving performance under attribute-wise imbalance scenarios. 
This indicates that their improvements in GLT protocols primarily stem from class-wise invariance.
Notably, Meta-LP also attains competitive performance on GLT datasets due to the use of balanced metadata spanning classes and attributes. Nevertheless, CLP outperforms Meta-LP by effectively countering the most common form of spuriousness, the correlation between backgrounds and labels, which positively impacts its performance.
The results also demonstrate that augmentation methods typically surpass long-tailed transfer learning methods in GLT protocols. In comparison to other augmentation approaches, our method solely augments the metadata, demonstrating greater efficiency and effectiveness.

\subsection{Visualization Experiments}
We utilize GradCAM for a visual examination to assess the impact of CLP on model performance, specifically by observing if the models prioritize foregrounds. The visualizations of the regions employed by the models for predictions are depicted in Fig.~\ref{visua}. These visualizations manifest that while the baseline model frequently directs attention towards non-relevant backgrounds or nuisances, leading to false predictions, our approach effectively centers on the causal regions corresponding to the objects, aiding the models in making correct classifications. Though some cases may still result in false predictions, as exemplified by the Rhodesian ridgeback, CLP significantly promotes model focus on image foregrounds. As a consequence, spurious correlations between background features and labels are mitigated.

\begin{figure}[t] 
\centering
\includegraphics[width=0.85\textwidth]{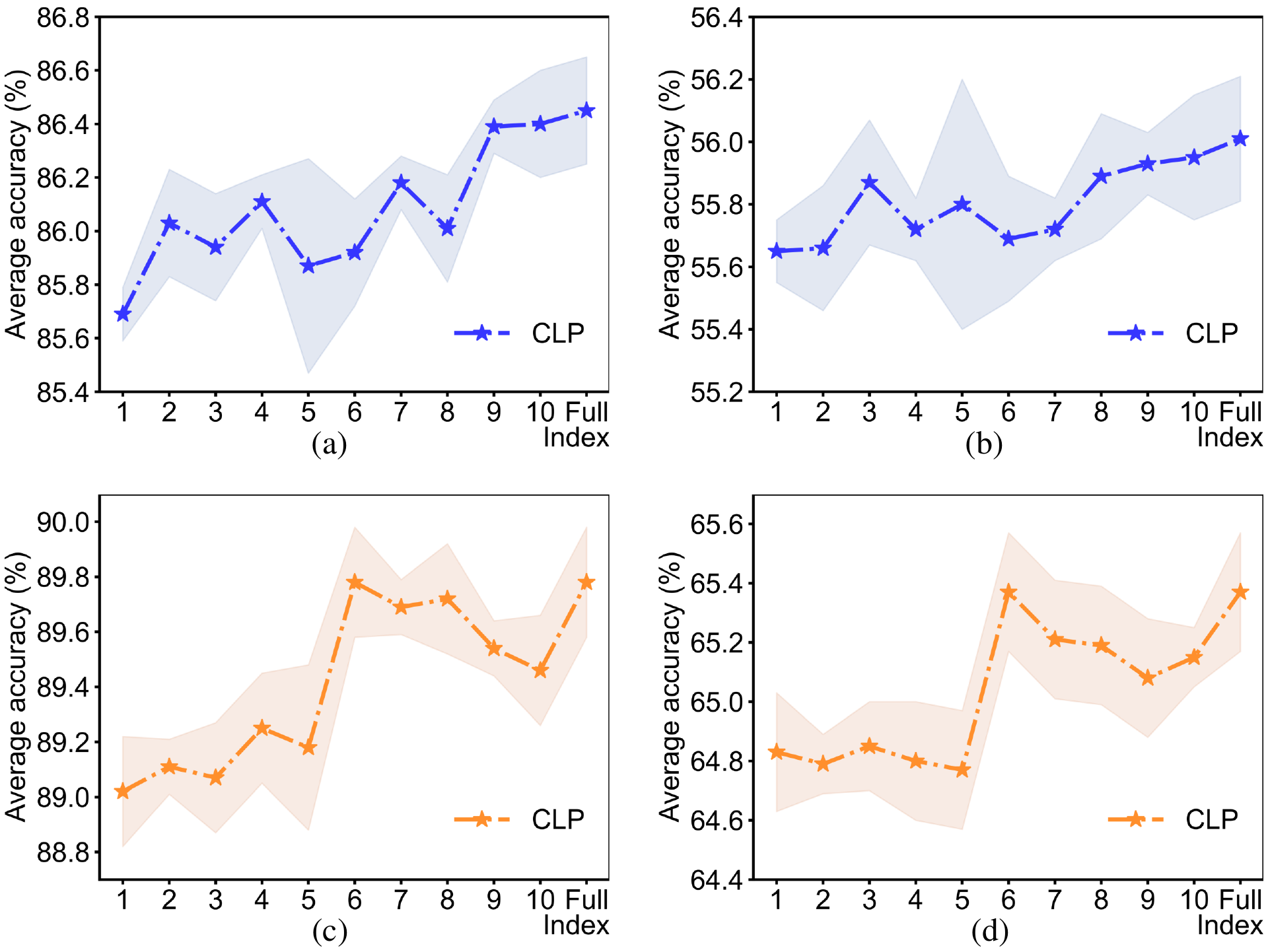}
\caption{{Ablation studies of extracted training characteristics under LT learning scenarios for CIFAR10 (a) and CIFAR100 (b) datasets. Ablation studies of extracted training characteristics under noisy label learning scenarios for CIFAR10 (c) and CIFAR100 (d) datasets.
Each data point indicates the performance of our CLP framework when the training feature corresponding to the index at the current horizontal coordinate is removed.}}
\label{aba}
\end{figure}
\subsection{Ablation and Sensitivity Studies}
In this section, we first investigate CLP with and without the saliency regularization term. These experiments are performed on the imbalanced CIFAR10 and CIFAR100 datasets, with imbalance ratios of 10:1 and 100:1, respectively. Optimal augmentation strategies tailored to each dataset are employed; for instance, CF(Tail)+F(Shuffle) is utilized on CIFAR10 with an imbalance factor of 10:1. The results, illustrated in Fig.~\ref{salaba}, underscore the crucial role and indispensability of the saliency regularization term.

Subsequently, we proceed to perform sensitivity tests to assess the impact of $\lambda$. The results on CIFAR data with 20\% flip noise, as illustrated in Figs.~\ref{sal2}(a) and (b), reveal the stable performance of CLP across a range of $\lambda\in[0.2,1]$. Furthermore, we conduct a sensitivity analysis for the number of augmented meta samples on noisy CIFAR data. The ratio of uniform noise is set to 40\%. The size of the original metadata is fixed at 1,000, and the number of samples augmented through both counterfactual and factual augmentations was maintained at a 1:1 ratio. The findings, depicted in Figs.~\ref{sal2}(c) and (d), indicate an improvement in model performance as the size of augmented data increases. However, once the ratio of augmented samples to the original metadata size exceeds 10:1, the performance stabilizes. 

{Finally, we perform ablation studies to assess the effectiveness of the extracted ten characteristics. These experiments are conducted on both LT learning and noisy label learning tasks. As shown in Fig.~\ref{aba}, the performance is optimal when all ten characteristics are included, thereby highlighting the efficacy of each characteristic.}


\section{Discussion}
This study primarily focuses on investigating a specific type of spuriousness, which involves distinguishing between spurious and causal features. This distinction allows us to utilize foreground annotations to address spurious associations between non-robust attributes and classes. However, it is important to note that our CLP framework can easily be extended to address other forms of spuriousness, such as those related to colors and shapes. This can be accomplished by enhancing the metadata with corresponding counterfactual and factual samples. For example, in situations where the color blue is mistakenly linked with a class, we can generate counterfactual images that showcase the blue color but with inaccurate foreground annotations, and factual images with random colors but accurate foreground annotations. Utilizing a meta dataset augmented in this manner to guide the training of the perturbation network can assist the model in effectively disentangling spurious associations between colors and categories.


\section{Conclusion}

This study introduces a novel logit perturbation framework (i.e., CLP) to train classifiers using the generated causal logit perturbations, which effectively mitigates spurious correlations between image backgrounds and labels. The entire framework is updated through a meta-learning-based algorithm. To incorporate human causal knowledge into the metadata, the meta dataset undergoes augmentation in both factual and counterfactual manners, which can then guide the training of the perturbation network. A series of training characteristics are extracted from the classifier and fed into the perturbation network to generate sample-wise logit perturbations. Extensive experiments are conducted across four typical learning scenarios influenced by spurious correlations, conclusively demonstrating the superior performance of CLP compared to other advanced approaches.





\end{document}